\documentclass[journal]{IEEEtran}
\IEEEoverridecommandlockouts

\usepackage{moreverb,url}
\usepackage[dvipsnames]{xcolor}
\usepackage{url} 
 
\usepackage{times} 
\usepackage{fixltx2e} 
\usepackage{multicol} 
\usepackage{graphicx} 
\usepackage[utf8]{inputenc} 
\usepackage{graphics}
\usepackage{epsfig} 
\usepackage{bm} 
\usepackage{caption} 
\usepackage{subcaption} 
\usepackage{ifthen} 
\newboolean{include-notes} 
\usepackage[linesnumbered,ruled,noend]{algorithm2e} 
\usepackage[noend]{algpseudocode} 
\usepackage{amssymb}
\usepackage{bm}
\usepackage[numbers]{natbib}
\usepackage[bookmarks=true]{hyperref}
\usepackage{amsmath}
\usepackage{mathabx}
\usepackage{multicol}
\usepackage{lipsum}  
\usepackage{dblfloatfix}

\DeclareMathOperator*{\argmin}{arg\,min} 
\renewcommand\vec{\mathbf} 

\usepackage[english]{babel}

\begin{document}

\title{Robust Physics-Based Manipulation by Interleaving Open and Closed-Loop Execution}

\author{Wisdom C. Agboh \hspace{20mm} Mehmet R. Dogar \vspace{3mm} \\
School of Computing \\ University of Leeds \\
\{w.c.agboh, m.r.dogar\}@leeds.ac.uk
}


\maketitle


\begin{abstract}
We present a planning and control framework for physics-based manipulation under uncertainty. The key idea is to interleave robust open-loop execution with closed-loop control.
We derive robustness metrics through contraction theory. We use these metrics to plan trajectories that are robust to both state uncertainty and model inaccuracies. However, fully robust trajectories are extremely difficult to find or may not exist for many multi-contact manipulation problems. We separate a trajectory into robust and non-robust segments through a minimum cost path search on a robustness graph. Robust segments are executed open-loop and non-robust segments are executed with model-predictive control. 
We conduct experiments on a real robotic system for reaching in clutter. Our results suggest that the open and closed-loop approach results in up to 35\% more real-world success compared to open-loop baselines and a 40\% reduction in execution time compared to model-predictive control. 
We show for the first time that partially open-loop manipulation plans generated with our approach reach similar success rates to model-predictive control,
while achieving a more fluent/real-time execution. A video showing real-robot executions can be found at \url{https://youtu.be/rPOPCwHfV4g}. 

\end{abstract}
\section{Introduction}
\label{sec:introduction}
The questions of how to achieve robotic manipulation success under uncertainty, and how to realize fluent/real-time manipulation plan execution remain. These problems become even more challenging for tasks where physics predictions play a central role \cite{Toussaint_RAL_2020, Saleem_ICRA_2020, Toussaint_RSS_2018}. Such a physics-based manipulation task can be seen in Fig. 1, where the goal is to reach for the green can without pushing other objects off the shelf. 

One traditional approach for such tasks is to generate a manipulation plan and execute the actions open-loop, without feedback \cite{RRT, PRM, Hassan_ICRA2020}. Such plans fail in the face of uncertainty – in state estimation and physics predictions \cite{Agboh_Humanoids2018}. Prior work has investigated the generation of robust open-loop manipulation plans \cite{pkpiece, Koval_IROS_2015, convergent_planning}. These robust open-loop plans are guaranteed to succeed in the face of uncertainty. They may include certain funneling actions that reduce uncertainty. Such completely open-loop robust plans are difficult to find, or may not exist for many physics-based manipulation problems.  

Another approach in prior work is to incorporate feedback, and re-plan online, during plan execution \cite{Agboh_Humanoids2018, Arruda_Humanoids_2017, Abraham_RAL_2020}. While such a closed-loop approach significantly improves task success, it is slow, due to the execution being interrupted repeatedly by computationally expensive planning/optimization cycles. This is particularly evident during physics-based manipulation, due to the large number of expensive physics-predictions required during planning.

In this work, we propose a planning and control framework that tightly integrates open and closed-loop execution for physics-based manipulation. It autonomously generates a combination of robust open-loop plans (wherever possible) and closed-loop controllers (wherever needed) to complete a manipulation task.

We seek robust open-loop plans wherever possible for three reasons. First, trajectory execution is fluent/real-time since the robot does not spend time re-planning at every step \cite{pkpiece, Koval_IROS_2015}. Second, robust open-loop plans imply near sensor-less manipulation, as opposed to full state estimation at every step, which may not be possible for some manipulation tasks \cite{Erdmann_Robotics_and_Automation_1988, Goldberg_Algorithmica_1993}.
Third, closed-loop model-predictive control with a high re-planning/re-optimization time suffers for dynamic manipulation tasks where objects may keep moving significantly during re-optimization whereas fluent/open-loop plans \cite{Haustein_ICRA_2015, dogar_clutter} can work better in such dynamic environments.

\begin{figure*}[htb!]
\centering
\vspace{10mm}
\begin{picture}(0,0)
    \put(50,13){\color{blue}{Closed-loop}}
    \put(57,3){\color{blue}{control}}
    \put(-102,13){\color{ForestGreen}{Robust}}
    \put(-107,3){\color{ForestGreen}{open-loop}}
    \put(153,13){\color{ForestGreen}{Robust}}
    \put(148,3){\color{ForestGreen}{open-loop}}
\end{picture}
\includegraphics[height=0.6\textwidth, width=\textwidth]{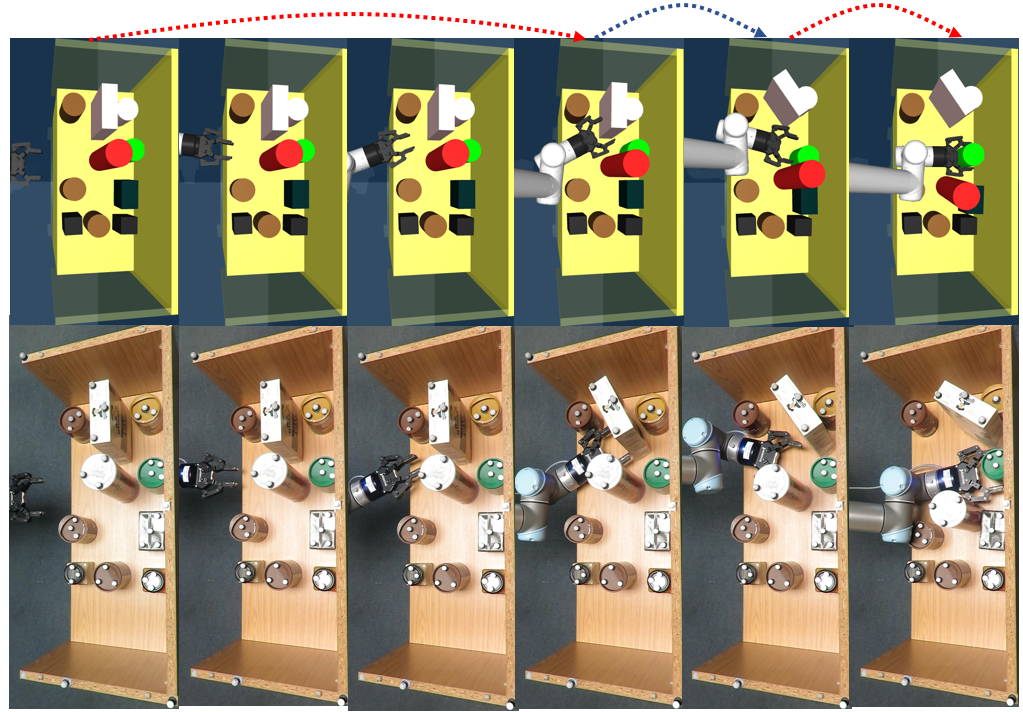}
\caption{A combination of robust open-loop execution and closed-loop control to reach for the green target object. Top row: Planned trajectories and execution strategy. Bottom row: Real-robot open and closed-loop execution to reach for the green target object. The robot starts with robust open-loop loop execution due to motion in free-space. It falls back to closed-loop control during multi-contact interactions. Finally, it uses a robust funnelling motion to get the green target object in the gripper.}
\label{fig:robot_fig_1}
\end{figure*}
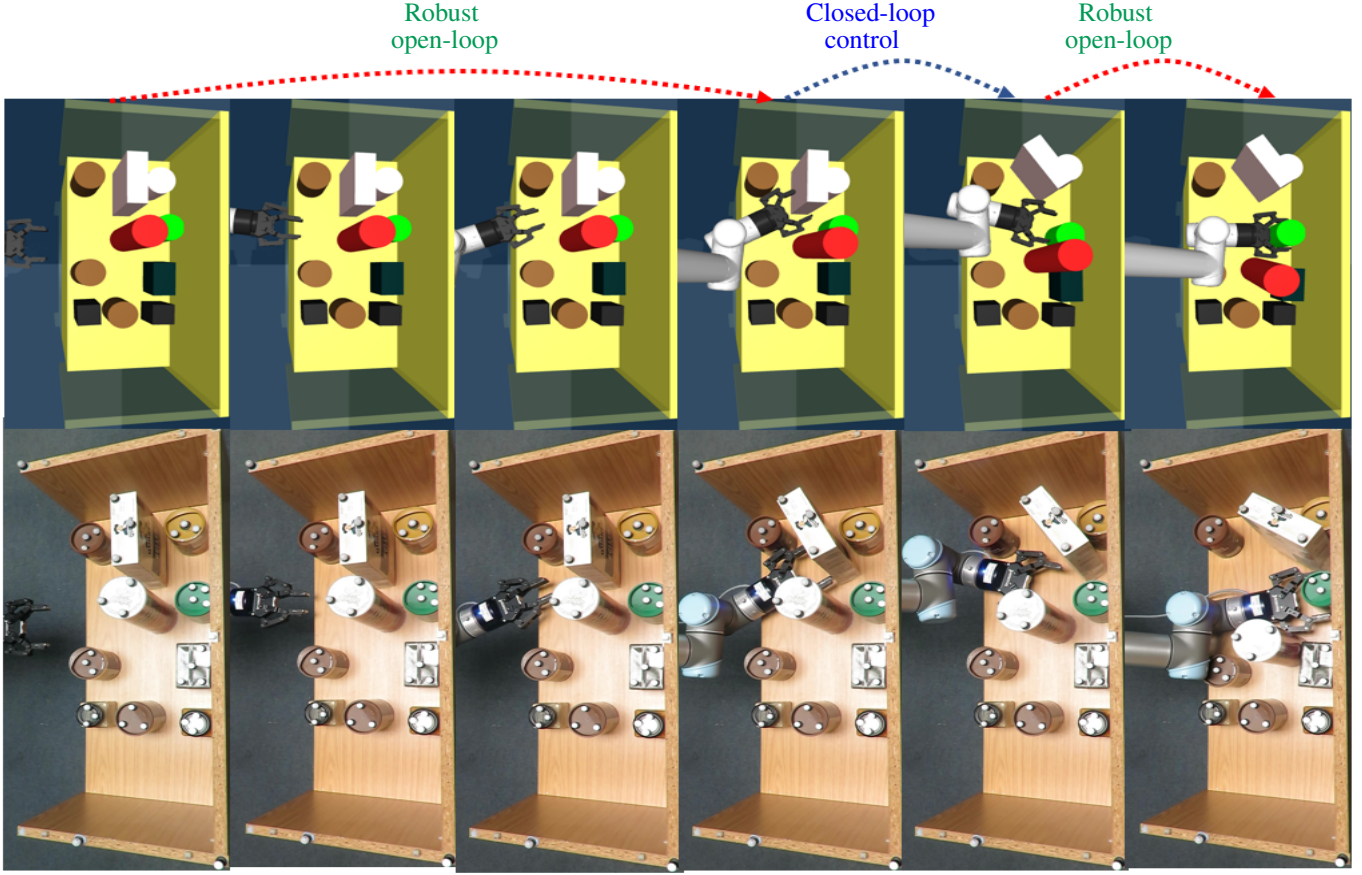

Consider the scene in Fig.~\ref{fig:robot_fig_1}. This is a scene that requires a large number of contact interactions. A robust open-loop plan that solves the complete task here is difficult to find due to uncertainty in state and more importantly in physics predictions. A closed-loop controller may be successful but will be slow due to computationally expensive physics predictions, re-planning at each step to reach the goal. 

Our approach generates a three-part plan for this scene. The first part is a robust open-loop plan. It is possible mainly due to the near free-space motion of the robot. The second part is executed closed-loop since there is a significant amount of contact interactions. The final part is also a robust open-loop plan. While there are still significant contact interactions, the final part involves a funnelling action that reduces uncertainty through object contact with the shelf. 

How do we detect what parts are robust open-loop and what parts should be closed-loop? We need a metric that quantifies robustness to state uncertainty and model inaccuracies.

We propose such a robustness metric in this work, based on contraction theory \cite{contraction_theory}. Prior work by \citet{convergent_planning} and \citet{johnson_isrr19} have proposed robustness metrics for state uncertainty. We build on these works, and derive new metrics to quantify robustness to not only state uncertainty, but more importantly model inaccuracies.  We use the metrics in a robust planner based on trajectory optimization. 

After robust planning for a given manipulation task, we analyze the plan to extract robust parts which are executed open-loop and non-robust parts which are executed with model-predictive control. We separate a plan into robust and non-robust segments through search on a directed graph, where nodes are time-points of a trajectory and edges are robust or non-robust connections between these time-points.

We find that the interleaved open and closed-loop execution approach leads to significant success under uncertainty with fluent/real-time execution, compared to baselines in the literature. 

More specifically, we show for the first time, that partially open-loop plans can produce similar success rates to full closed-loop control during physics-based manipulation.

We make the following contributions: 

\begin{itemize}
    \item A planning and control framework that autonomously switches between robust open-loop execution, and model-predictive control to complete a physics-based manipulation task. 
    
    \item A derivation of divergence metrics through contraction theory, to quantify robustness to state uncertainty, in the presence of real-world model inaccuracies. 
    
    \item A novel robust planner based on trajectory optimization. 
\end{itemize}

The rest of this paper is organized as follows: In Sec.~\ref{sec:related_work} we discuss related work in the literature. Sec.~\ref{sec:problem_statement} formulates the problem. Sec.~\ref{sec:overview} provides an overview of the interleaved open and closed-loop execution approach. Sec.~\ref{sec:divergence_metrics} provides a background on contraction theory and our robustness metric derivation. Sec.~\ref{sec:robust_planning} introduces the robust planning and control framework. Sec.~\ref{sec:experiments} details robot experiments and results. Sec.~\ref{sec:conclusion} concludes the paper. 

\section{Related work}
\label{sec:related_work}

Previous work considers physics-based manipulation either as an open-loop planning problem or a closed-loop control problem. Within open-loop planning, there have also been attempts to generate robust plans. We review these three approaches below.

\subsection{Open-loop execution} 
Prior work has presented planners to address the physics-based manipulation problem, with open-loop execution on a real robot. \citet{kitaev_abbeel} propose a physics-based trajectory optimizer to generate manipulation plans to reach into cluttered environments. \citet{king2015nonprehensile} present a randomized kinodynamic planner to address the rearrangement planning problem. \citet{dogar_clutter} present a planning method for grasping in clutter where multiple robot-object interactions are possible. \citet{Zito_IROS2012} present a two-level RRT planner one for the global path, and the other to plan local pushes. \citet{Cruciani_IROS_2017} proposed a three-stage method for in-hand manipulation, with a focus on pivoting.

While these open-loop methods succeed in limited settings, their success rates decrease in more complex uncertain environments with increased multi-contact interactions \cite{Agboh_Humanoids2018}.  

\subsection{Robust open-loop execution}
To increase success rates during open-loop execution, prior work has considered uncertainty at the planning stage. 

\citet{Anders_ICRA_2018} take a conformant planning approach to generate planar robust push actions through a learned belief-state transition model. \citet{pkpiece} propose p-KPIECE, a randomized physics-based planner that accounts for object pose and contact dynamics uncertainty. \citet{convergent_planning} use contraction theory to derive divergence metrics that are then used to generate robust manipulation plans. \citet{Koval_IROS_2015} formulate robust open-loop planning as a multi-armed bandit problem, and pick the best arm in a rearrangement task. \citet{cc_RRT} propose chance constrained rapidly-exploring random trees. It handles environmental uncertainty by considering the trade-off between planner conservatism and the risk of infeasibility. \citet{Zhou_RAL_2017} propose a probabilistic algorithm to sequentially reduce state uncertainty during grasping until an object's pose is uniquely known, under stochastic dynamics. 

With these robust open-loop planners, success rates increase compared to traditional open-loop execution. However, completely robust open-loop plans are difficult to find as they may depend on a particular set of "funneling" actions. These completely robust plans tend to be pessimistic/conservative and may not exist for many physics-based manipulation problems.

\subsection{Closed-loop control policies}

Feedback during execution can help improve success rates in physics-based manipulation.

\citet{Abraham_RAL_2020} extend path integral control by embedding uncertainty in the action, to provide robust manipulation through online control. \citet{Hogan_ICRA_2018} search for contact mode sequences offline, and optimal control inputs online in a convex hybrid MPC setting, for reactive planar contact-based manipulation. \citet{Papallas_ICRA_2020} include human-operator input during closed-loop execution to improve success rates in physics-based manipulation.  \citet{Arruda_Humanoids_2017} learn forward dynamics models, and propose an uncertainty-averse planner based on path integral control, to push objects away from high uncertainty regions. \citet{Huang_ICRA_2019} solve a tabletop rearrangement planning problem with policy roll-outs and an iterated local search approach, in an online setting. \citet{Agboh_WAFR2018} provide an online solution to the physics-based manipulation problem. It embraces uncertainty during pushing, to generate fast or slow robot pushes depending on the task. 

While these prior methods have achieved improved success rates under uncertainty, a major drawback in online control here is computationally expensive physics predictions that lead to high re-planning times \cite{Agboh_CVS2020, Agboh_ISRR2019}.

Robust learned robot policies have also been developed in prior work for physics-based manipulation. \citet{Bejjani_Humanoids_2018} use a learned value function and look-ahead planning, in an online setting, to generate physics-based manipulation plans in clutter. \citet{Yuan_RAS_2019} use deep reinforcement learning to learn an object rearrangement policy which is then further adapted for the real-world through additional real-world manipulation data. \citet{Laskey_CASE_2016} learn grasping in clutter from demonstrations provided by a hierarchy of supervisors, to reduce the burden on human experts to provide demonstrations. \citet{Pauly_arxiv_2020} learn to perform robot manipulation tasks from a single third-person demonstration video. \citet{Li_RSS_2018} propose push-net, a deep recurrent neural network that uses only an image as input to push objects of unknown physical properties. \citet{Bejjani_IROS_2021} address the problem of occlusion in lateral access into shelves, with a hybrid planner based on a learned heuristic. \citet{Kiatos_ICRA_2019} learn an optimal push policy to singulate objects in clutter with lateral pushing actions.  

These learned policies have shown impressive success rates with typically low policy-lags for real-world manipulation tasks. However, they require a large amount of training data/demonstrations for a given task, and do not generalize well to new tasks. 

This work proposes a planning and control framework that gets the best of both robust open-loop execution, and closed-loop control. It generates robust open-loop actions wherever possible, and reverts to model-predictive control otherwise, to complete a physics-based manipulation task. Thus, it achieves significant success under uncertainty with fluent/real-time execution. 

\section{Problem statement}
\label{sec:problem_statement}
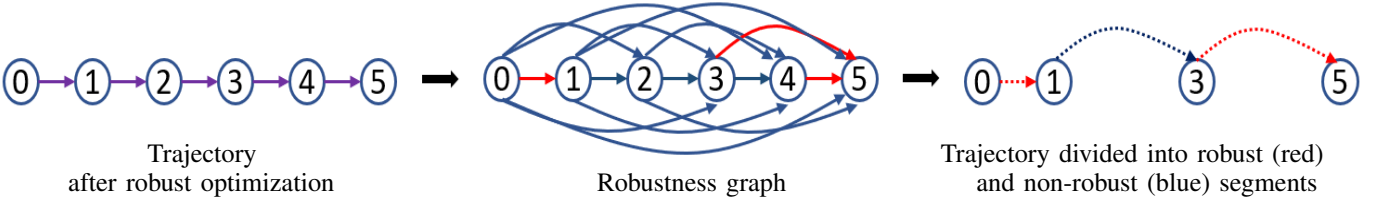
\begin{figure*}[htb!]
\centering
\includegraphics[height=0.12\textwidth, width=\textwidth]{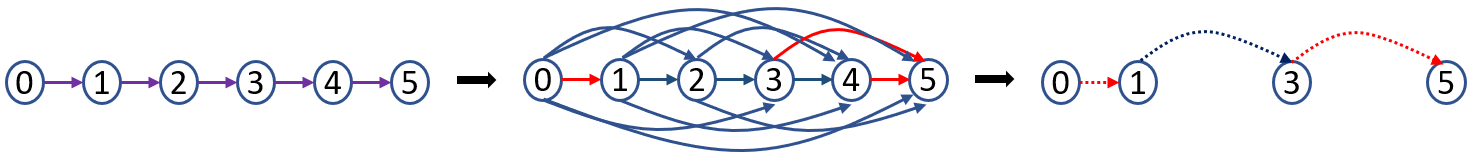}
\begin{picture}(0,0)
    \put(-200, 11){Trajectory}
    \put(-230, 0){after robust optimization}
    \put(-30, 0){Robustness graph}
    \put( 100, 11){Trajectory divided into robust (red) }
    \put(110, 0) { and non-robust (blue) segments}
\end{picture}
\vspace{3mm}
\caption{Search in a robustness graph to find a combination of robust and non-robust segments that maximize the number of time-points that are robust. In the robustness graph, the robust edges are shown in green, the non-robust edges are shown in blue.}
\label{fig:robustness_graph}
\end{figure*}

In this work, we focus on an example task where the robot's goal is to retrieve an item from a cluttered environment under state uncertainty and model inaccuracies. It starts from an observed initial state and generates non-prehensile actions to reach a final pre-grasping state. 

A scene includes at most $D$ movable dynamic objects. $\vec{q}^{i}, \hspace{2mm} i=1,\dots,D$, refers to the full pose of each dynamic object. The robot's pose is defined by a vector of joint values $\vec{q}^{R}$. We represent the complete state of our system as $\vec{x}_{t}  \in \mathbb{R}^{n}$ at time $t$. This includes the pose and velocity of the robot and all dynamic objects; $\vec{x}_{t} = \{\vec{q}^{R}, \vec{q}^{1}, \dots, \vec{q}^{D},\dot{\vec{q}}^{R},\dot{\vec{q}}^{1}, \dots, \dot{\vec{q}}^{D}\}$. 

The control inputs, $\vec{u}_{t} \in \mathbb{R}^{m}$ are velocities applied to the robot's joints: $\vec{u}_{t}=\dot{\vec{q}}^{R}$ for a fixed control duration $\Delta_{t}$. Then, the discrete time dynamics  of the system is:
\begin{equation}\label{eq:dt_dynamics}  
 \vec{x}_{t+1} = f(\vec{x}_{t}, \vec{u}_t)
\end{equation}
where $f$ : $\mathbb{R}^{n} \times \mathbb{R}^m \longmapsto \mathbb{R}^{n}$ is the state transition function and we assume an observed initial state of the system, $\vec{x}_0$. 

System dynamics in Eq.~\ref{eq:dt_dynamics} is modeled with the physics engine Mujoco \citep{mujoco}. Nevertheless, any physics engine is an inaccurate model of the real-world physics and uncertainties over the system dynamics are inevitable. Indeed, even if we assumed perfect modelling, it is difficult for a robot to know the exact geometric, frictional, and inertial properties of objects in an environment. Thus, we represent the real-world dynamics as:
\begin{equation} \label{eq:dt_dynamics_disturbance}
    \vec{x}_{t+1} = f(\vec{x}_{t} , \vec{u}_{t}) + w(\vec{x}_{t}, \vec{u}_{t}) 
\end{equation}
where $w(\vec{x}_{t}, \vec{u}_{t}): \mathbb{R}^{n} \times \mathbb{R}^m \longmapsto \mathbb{R}^{n}$ is an unknown disturbance term  due to model inaccuracies. 

Our goal in this paper is to build a robust planning and control framework for physics-based manipulation in clutter. The robot must generate control inputs, $\vec{u}_{t}$ at time $t$ that drive the system in Eq.~\ref{eq:dt_dynamics_disturbance}, from an observed initial state $\vec{x}_{0}$, to a desired goal state set, under state uncertainty and model inaccuracies.

In the foregoing paragraphs, $\vec{U} = [\vec{u}_{t_{0}},\vec{u}_{t_{1}}, \dots, \vec{u}_{t_{N-1}}]$ denotes a sequence of control inputs of length $N$. Similarly, a sequence of states is  $\vec{X} = [\vec{x}_{t_{0}},\vec{x}_{t_{1}}, \dots, \vec{x}_{t_{N}} ]$. 

\section{Overview}
\label{sec:overview}

In this section, we provide our overall framework called $OCL$, for \textit{open and closed-loop execution} during physics-based robotic manipulation. Our target is open-loop trajectory execution in the real-world wherever possible. We aim to find robust plans that are guaranteed to be \textit{successful} under state uncertainty and model inaccuracies. However, this is not always possible. Therefore, when the open-loop trajectory is not guaranteed to be successful, we design and use a feedback controller.

$OCL$ is presented in Alg.~\ref{alg:OCL}.
\setlength{\textfloatsep}{2mm}
\begin{algorithm}[t]
    \SetKwInOut{Input}{Input}
    \SetKwInOut{Output}{Output}
    \SetKwInOut{Parameters}{Parameters}
    \SetKwInOut{Subroutines}{Subroutines}
    \Input{$\vec{x}_{0}$: Initial state \\ 
     $\vec{U}$: Initial candidate control sequence}
     $\vec{X}^{*}, \vec{U}^{*}, \hat{\vec{E}}^{r}_{e} \gets$ RobustSTO $(\vec{x}_{0},  \vec{U})$ \\
     RobustSegs, \hspace{0.2mm} NonRobustSegs $\gets$ GetSegments$(\hat{\vec{E}}^{r}_{e})$ \\
     %
    \For{\hspace{0.1mm} seg \hspace{0.1mm} $\mathbf{in}$ \hspace{0.1mm} $\mathrm{RobustSegs}$ $\cup$ $\mathrm{NonRobustSegs}$ \hspace{0.1mm}}{
    \If{seg $\mathbf{in}$ $\mathrm{RobustSegs}$}
    {Execute $\vec{U}^{*}_{seg}$ open-loop}
    \Else{Execute $\vec{U}^{*}_{seg}$ with MPC}}
    \caption{Open and Closed-loop (OCL)}\label{alg:OCL}
\end{algorithm}
\setlength{\floatsep}{2mm}

It begins with robust planning by minimizing a robust objective function with trajectory optimization (line 1). We derive real-world divergence metrics in Sec.~\ref{sec:divergence_metrics}. They quantify robustness to both state uncertainty and model inaccuracies. We use these metrics in the robust objective. Details of the robust planner can be found in Sec.\ref{sec:robust_planning}. The robust planner also returns the corresponding robustness metric ($\hat{\vec{E}}^{r}_{e}$) for the planned trajectory.  

Robust planning does not always find a trajectory that is guaranteed to be open-loop robust from start to end. However, some parts of this trajectory may be robust. Hence we seek to divide a given trajectory into a combination of robust and non-robust segments with the GetSegments(.) subroutine (line 2). Thereafter, the robust segments are executed open-loop while the non-robust segments are executed with model-predictive control (lines 3-7).

The GetSegments(.) subroutine generates a directed graph to address the problem of dividing the trajectory ($\vec{X}^{*}$, $\vec{U}^{*}$) into robust and non-robust parts. The nodes of this robustness graph are all the time-points of the trajectory and edges are connections between these nodes, forward in time. An edge is robust or non-robust, indicating that the trajectory segment between those two time-points is robust or non-robust. Details of this segment robustness metric computation can be found in Sec.~\ref{sec:edge_metric_computation}.
To find a complete plan with as many robust segments as possible, we convert the problem into a graph search by assigning costs to each edge. A robust edge going from node $i$ to $j$ costs $\frac{c_{ro}}{j-i}$, and a non-robust edge costs $c_{nr} \cdot (j-i)$, where $c_{ro}<<c_{nr}$, and are constants, and $i<j$. We search this robustness graph for the lowest cost path from the start point 0, to the end point of the trajectory, N. The output is a path consisting of robust and non-robust segments. 

Consider the example in Fig.~\ref{fig:robustness_graph}. On the left, we have a trajectory with 6 time-points, the output of robust planning. We build a robustness graph with robustness metrics for the trajectory (center), where robust segments are in red and non-robust segments are in blue. We search this graph for the lowest cost path from time-point 0 to time-point 5. It produces a robust segment (0-1), a non-robust segment (1-3), and a robust segment (3-5). The robust segments are executed open-loop and the non-robust segment is executed with model-predictive control. 


\section{Robustness to uncertainty} 
\label{sec:divergence_metrics}

We define divergence metrics through contraction analysis to quantify robustness to state uncertainty for a nominal trajectory, under model inaccuracies. We provide a brief introduction to contraction analysis, a derivation of divergence metrics for the real-world case, and a corresponding numerical approximation. 

\subsection{Contraction analysis:} 
Contraction theory \cite{contraction_theory} studies the evolution
of infinitesimal distance between any two neighboring trajectories and provides conclusions on the finite distance between them. 

We begin with the continuous time version of the autonomous system in Eq.~\ref{eq:dt_dynamics}:

\begin{equation}\label{eq:ct_dynamics}
    \dot{\vec{x}} = f(\vec{x}(t), \vec{u}(t)) 
\end{equation}

Consider two neighbouring trajectories separated by a virtual displacement $\delta{\vec{x}}$. The squared distance between them is $\delta{\vec{x}^{T} \delta{\vec{x}}}$. The rate of change of this distance is given by:

\begin{equation}\label{eq:rate_of_change}
\frac{d}{dt} (\delta{\vec{x}^{T} \delta{\vec{x}}}) = 2\delta{\vec{x}^{T} {\delta{\dot{\vec{x}}}}}
\end{equation}

From Eq.~\ref{eq:ct_dynamics}, ${\delta{\dot{\vec{x}}}} = \frac{\partial{f}}{\partial{\vec{x}}} \delta{\vec{x}}$. Therefore Eq.~\ref{eq:rate_of_change} becomes:
\begin{equation}\label{eq:rate_of_change_modified}
    \frac{d}{dt} (\delta{\vec{x}^{T} \delta{\vec{x}}}) = 2\delta{\vec{x}^{T}} \frac{\partial{f}}{\partial{\vec{x}}} \delta{\vec{x}} 
\end{equation}
The Jacobian of $f$, $\frac{\partial{f}}{\partial{\vec{x}}}$ can be written as a sum of  symmetric and skew-symmetric parts:
\begin{equation}\label{eq:f_jacobian}
    \frac{\partial{f}}{\partial{\vec{x}}} = \frac{1}{2} ( \frac{\partial{f}}{\partial{\vec{x}}} +  \frac{\partial{f}^{T}}{\partial{\vec{x}}}) + \frac{1}{2} ( \frac{\partial{f}}{\partial{\vec{x}}} -  \frac{\partial{f}^{T}}{\partial{\vec{x}}})
\end{equation}
Let $\lambda{^{f}_{max}}(\vec{x}, \vec{u})$  be the largest eigenvalue of the symmetric part (first term of Eq.~\ref{eq:f_jacobian}) of the Jacobian $\frac{\partial {f}}{\partial{\vec{x}}}$. Then, given that the eigenvalue of the skew-symmetric part is 0 or imaginary,  it follows that:
\begin{equation}
        \frac{d}{dt} (\delta{\vec{x}^{T} \delta{\vec{x}}}) \leq 2\lambda{^{f}_{max}}\delta{\vec{x}^{T}} \delta{\vec{x}} 
\end{equation}
If we define $ D^{f}_{m} := \lambda{^{f}_{max}}$ as the maximal divergence metric, we find:
\begin{equation}\label{eq:f_exp_convergence}
    || \delta{\vec{x}(t)} || \leq || \delta{\vec{x}(t_{0})} || e^{\int^{t}_{t_{0}} D^{f}_m(\vec{x}, \vec{u}, t) dt}
\end{equation}
From Eq.~\ref{eq:f_exp_convergence}, we find that if $D^{f}_{m}$ is uniformly negative definite, any infinitesimal distance $||\delta_{x}(t)||$ converges exponentially to zero. Moreover, any finite path converges exponentially to zero (by path integration). 

We reach an important conclusion of contraction analysis: Given a nominal trajectory $\Bar{\vec{x}}(t)$, a solution to Eq.~\ref{eq:ct_dynamics} with nominal controls $\bar{\vec{u}}(t)$, any other trajectory that starts in a region centered on $\bar{\vec{x}}(t)$ where $D^{f}_{m}$ is uniformly negative definite everywhere, converges exponentially to the nominal $\bar{\vec{x}}(t)$ (~\cite{contraction_theory}, Theorem 1). 

The system in Eq.~\ref{eq:ct_dynamics} models the real-world but can be inaccurate. Thus, we ask the following questions: If the nominal controls $\bar{\vec{u}}(t)$ are executed in the real-world, will the real trajectory $\bar{\vec{x}}^{r}(t)$ be different? More importantly, will the convergence properties for the real trajectory change? 


\subsection{Contraction analysis for the real-world}

Extending beyond contraction analysis used in prior work \cite{convergent_planning, johnson_isrr19}, we consider the continuous-time version of Eq.~\ref{eq:dt_dynamics_disturbance}, a representation of the real-world dynamics:
\begin{equation}\label{eq:ct_dynamics_w}
    \dot{\vec{x}} = f(\vec{x}(t), \vec{u}(t)) + w(\vec{x}(t), \vec{u}(t))
\end{equation}
%

Indeed, the resulting real nominal trajectory $\bar{\vec{x}}^{r}(t)$ may be different due to the disturbance $w$. More importantly, we show that the resulting convergence properties may change.

The real maximal divergence metric $D^{r}_{m}$ for the system in Eq.~\ref{eq:ct_dynamics_w} can be written as: 
\begin{equation}
    D^{r}_{m} := D^{f+w}_{m}(\vec{x}, \vec{u})
\end{equation}
where $D^{f+w}_{m} := \lambda{^{f+w}_{max}}(\vec{x}, \vec{u})$ is the largest eigenvalue of the symmetric part of the Jacobian $\frac{\partial {(f+w)}}{\partial{\vec{x}}}$. We reach this conclusion directly by applying steps in Eq.[\ref{eq:rate_of_change}-\ref{eq:f_exp_convergence}] to the real-world system in Eq.~\ref{eq:ct_dynamics_w}. 


Another important divergence metric we consider in this work is the expected divergence metric. It was first introduced by \citet{convergent_planning}. It quantifies the evolution of the expected value of a virtual displacement. From Eq.~\ref{eq:f_exp_convergence}, the expected divergence metric $D^{f}_{e}$ for the system in Eq.~\ref{eq:ct_dynamics} can be written as:
\begin{equation}\label{eq:f_exp_convergence_expectation}
    E[|| \delta{\vec{x}(t)} ||] = E[|| \delta{\vec{x}(t_{0})} ||]e^{\int^{t}_{t_{0}} D^{f}_e(\vec{x}, \vec{u}, \tau) d\tau}
\end{equation}
\begin{equation}\label{eq:de_definition}
 D^{f}_{e} := \frac{d}{dt} \ln E[|| \delta{\vec{x}}(t) ||]   
\end{equation}

For the real-world case, similar to the real maximal divergence metric, one can write the real expected divergence metric for the system in Eq.~\ref{eq:ct_dynamics_w} as:  
\begin{equation}
    D^{r}_{e} := D^{f+w}_{e}
\end{equation}
Consider now a state trajectory from an initial time $t_{0}$ to a final time $t_{N}$, and define divergence path metrics such that:
\begin{equation}\label{eq:path_metric}
    E^{f}_{h} = e^{\int^{t_{N}}_{t_{0}} D^{f}_{h} dt}, \hspace{1mm} h \in \{m,e\}
\end{equation}
The nominal trajectory starting from $t_{0}$ and ending at $t_{N}$ is convergent if $E^{f}_{h} < 1$. Recall that we are also interested in the real nominal trajectory $\bar{\vec{x}}^{r}$. Thus, we write: 
\begin{equation}
    E^{r}_{h} = e^{\int^{t_{N}}_{t_{0}} D^{f+w}_{h} dt}, \hspace{1mm} h \in \{m,e\}
\end{equation}
and the real nominal trajectory is convergent if $E^{r}_{h} < 1$, depending on the metric of choice - expected or maximal. 

\subsection{Metric approximations} \label{sec:numerical_approx}
\citet{convergent_planning} approximate divergence metrics $D^{f}_{m}$ and $D^{f}_{e}$ by approximating the virtual displacement $\delta{{\vec{x}}}$ with finite samples. Let $\delta{\vec{x}}(t) = \vec{x}^{i}(t) - \bar{\vec{x}}(t)$, where $\vec{x}^{i}(t)$ is a solution for sample $i$ to the same system (Eq.~\ref{eq:ct_dynamics}) for nominal controls $\bar{\vec{u}}(t)$, starting with a different initial condition:  $\delta{\vec{x}}(t_{0}) = \vec{x}^{i}(t_{0}) - \bar{\vec{x}}(t_{0})$. Then, from Eq.~\ref{eq:f_exp_convergence} we can write an approximation $\hat{D}^{f}_{m}$ to the maximal divergence metric for a small time step $\delta{t}$ as:
\begin{equation}\label{eq:approx_dm}
    \hat{D}^{f}_{m} := \frac{1}{\delta{t}} \ln \max_{i} \frac{|| \vec{x}^{i}(t+\delta{t}) - \bar{\vec{x}}(t+\delta{t}) ||}{ || \vec{x}^{i}(t) - \bar{\vec{x}}(t) || }
\end{equation}
where $i = 1,\dots,N_c$, and $N_{c}$ is the number of finite samples.

Similarly, from Eq.~\ref{eq:f_exp_convergence_expectation}, we can write an approximation  $\hat{D}^{f}_{e}$ to the expected divergence metric as:
\begin{equation}\label{eq:approx_de}
    \hat{D}^{f}_{e} := \frac{1}{\delta{t}} \ln \frac{ \frac{1}{N_c} \sum^{N_{c}}_{i=0} || \vec{x}^{i}(t+\delta{t}) - \bar{\vec{x}}(t+\delta{t}) ||}{ \frac{1}{N_c} \sum^{N_{c}}_{i=0} || \vec{x}^{i}(t) - \bar{\vec{x}}(t) || }
\end{equation}

From Eq.~\ref{eq:approx_dm} and Eq.~\ref{eq:approx_de}, and considering the definition in Eq.~\ref{eq:path_metric}, we define approximations $\hat{E}^{f}_{e}$ to the expected divergence path metric and  $\hat{E}^{f}_{m}$ to the maximal divergence path metric valid for a finite time from $t_{0}$ to $t_{N}$: 
\begin{equation}
    \hat{E}^{f}_{e} :=  \frac{ \frac{1}{N_c} \sum^{N_{c}}_{i=0} || \vec{x}^{i}(t_{N}) - \bar{\vec{x}}(t_{N}) ||}{\frac{1}{N_c} \sum^{N_{c}}_{i=0} || \vec{x}^{i}(t_{0}) - \bar{\vec{x}}(t_{0}) || }
\end{equation}

\begin{equation}
    \hat{E}^{f}_{m} :=  \max_{i} \frac{ || \vec{x}^{i}(t_{N}) - \bar{\vec{x}}(t_{N}) ||}{|| \vec{x}^{i}(t_{0}) - \bar{\vec{x}}(t_{0}) ||}, \hspace{1mm}  i = 0,1,\dots,N_c 
\end{equation}

If $\hat{E}^{f}_{h} < 1, \hspace{1mm} h \in \{m,e\}$, the nominal trajectory $\bar{\vec{x}}(t)$ from $t_0$ to $t_{N}$ is convergent, depending on one's choice of robustness metric - maximal (m) or expected (e). 

\subsection{Metric approximations for the real-world}
\label{sec:segment_metrics}
\begin{figure}[t]
   \includegraphics[scale=0.6]{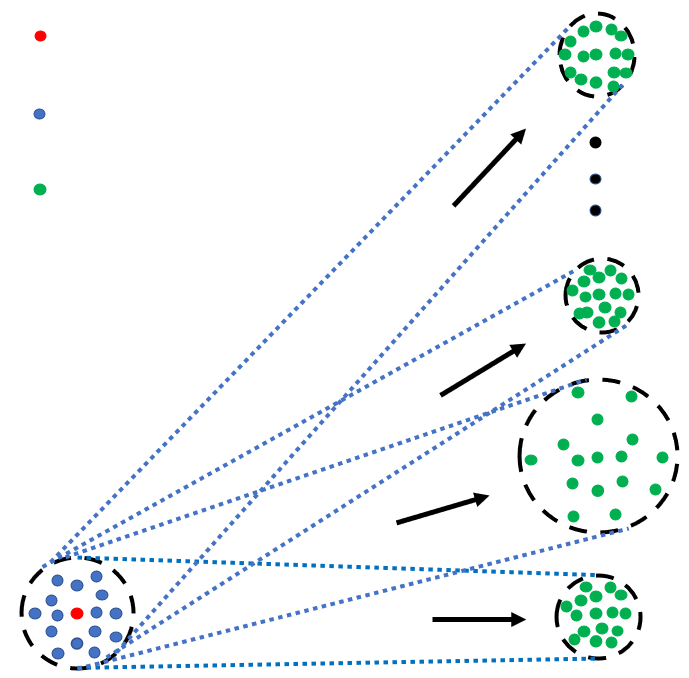}
   \begin{picture}(0,0)
   \put(-190,192){$\vec{x}_{0}$: Observed initial state}
    \put(-190,169){$\vec{x}^{i}_{0}$: Initial state sample i}
    \put(-190,146){$\vec{x}^{i}_{N}$: Final state sample i}
    \put(-2,15){$\hat{E}^{f}_{e} < 1$}
    \put(-2,65){$\hat{E}^{1}_{e} > 1$}
    \put(-2,112){$\hat{E}^{2}_{e} < 1$}
    \put(-9,186){$\hat{E}^{N_{w}}_{e} < 1$}
\end{picture}
  \caption{Computing the real-world expected divergence metric $\hat{E}^{r}_{e}$. First, we compute the divergence metric using the nominal model $f$ ($\hat{E}^{f}_{e}$). We draw $N_c$ sample initial states from the observed initial state $\vec{x}_{0}$ and roll-out a trajectory from each state, to obtain final state samples $\vec{x}^{i}_{N}$. Since the size of the final state distribution is less than that of the initial state distribution, $\hat{E}^{f}_{e} < 1$, and the trajectory is convergent in $f$.
  Thereafter, we randomly create $N_w$ real-world realizations from $f$ and compute the divergence metric in each of these $N_w$ worlds. These metrics can be very different from each other across the real-world realizations. We pick the metric with the maximum value as a worst-case approximation of the real-world divergence metric.}
  \label{fig:metric_computation}
 \end{figure}
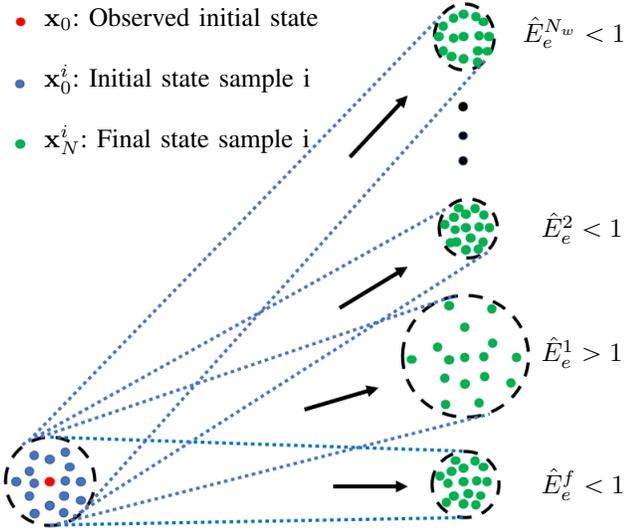

We define approximations to the real-world maximal and expected path divergence metrics:
\begin{equation}
    \hat{E}^{r}_{h} = \hat{E}^{f+w}_{h}, \hspace{1mm} h \in \{m,e\}
\end{equation}
The real-world nominal trajectory is convergent if $\hat{E}^{r}_{h} < 1 $. 

Given that the disturbance term $w(\vec{x}, \vec{u})$ is unknown, an important question is how to compute  $\hat{E}^{r}_{h}$. 

Consider a set of possible deterministic real-worlds:
\begin{equation}\label{eq:ct_dynamics_w_all}
    \dot{\vec{x}} = f(\vec{x}(t), \vec{u}(t)) + w_{j}(\vec{x}(t), \vec{u}(t)), \hspace{2mm} j=0,\dots,N_{w}
\end{equation}
where $N_{w} > 0$ is a finite number.

We assume that model inaccuracies resulting in $w_{j}$ arise only from inaccurate physics parameters. Specifically, we assume these parameters are the object's mass $m$, coefficient of friction $\bm{\mu}$, and size $\vec{p}$. These physics parameters are bounded such that:
\begin{equation}\label{eq:mass_bound}
    m_{l} \leq m \leq m_{u}, \hspace{1mm} \mu_{l} \leq \mu \leq \mu_{u},  \hspace{1mm} \vec{p}_{l} \leq \vec{p} \leq \vec{p}_{u}
\end{equation}
where $m_l$, $m_{u}$, $\mu_{l}, \mu_{u}$, are constants and $\vec{p}_{l}, \vec{p}_{u}$, are constant vectors. The size here is a vector of values describing the geometry of an object. For example, radius and height for a cylinder or length, breadth, and height for a box.  

Then, a sample real-world realization corresponds to running a physics engine that models $f$ with a uniformly sampled set of physics parameters. 

We consider a worst-case scenario and define $\hat{E}^{r}_{h}$ as the maximum divergence metric over all $N_w$ real-world realizations:

\begin{equation}
    \hat{E}^{r}_{h} := \max_{j} ( \hat{E}^{j}_{h}), \hspace{1mm} j = 0,\dots, N_{w}
\end{equation}
We provide an illustration in Fig.~\ref{fig:metric_computation}, for the expected real-world divergence metric.

\subsection{Segment metric computation} \label{sec:edge_metric_computation} 

Consider the discrete time points of a nominal trajectory $\bar{\vec{x}}_{t}$, where $t = t_0, t_1, \dots, t_{N}$. Then, the expected path metric for a one-step trajectory segment from $t_{p}$ to $t_{p+1}$ is: 
\begin{equation}\label{eq:one_step_segment}
\hat{E}_{e, t_{p}} = \frac{\frac{1}{N_c} \sum^{N_{c}}_{i=0} || \vec{x}^{i}_{t_{p+1}} - \bar{\vec{x}}_{t_{p+1}} ||}{\frac{1}{N_c} \sum^{N_{c}}_{i=0} || \vec{x}^{i}_{t_p} - \bar{\vec{x}}_{t_p} ||}
\end{equation}

Thus, we can write the divergence metric vector as:
\begin{align}
    \hat{\vec{E}}_{e} = [\hat{E}_{e,{t_{0}}}, \hat{E}_{e,{t_{1}}}, \dots, \hat{E}_{e,{t_{N}}}]
\end{align}

It is a list of one-step trajectory segment divergence metrics. Then, it follows from Eq.~\ref{eq:one_step_segment} that the divergence metric for any trajectory segment from time $t_{p}$ to time $t_{q}$ can be written as the following algebraic computation over $\hat{\vec{E}}_{e}$:

\begin{equation}
    \hat{E}_{e, t_{p}, t_{q}}  := \prod^{t=t_q}_{t=t_{p}} \hat{\vec{E}}_{e}[t]   
\end{equation}

Hence, a trajectory segment starting at $t_p$ and ending at $t_{q}$ is robust if $\hat{E}_{e, t_{p}, t_{q}} < 1$. Also, the divergence metric for the full trajectory can be written as:

\begin{equation}\label{eq:algebraic_computation}
    \hat{E}_{e}  = \prod^{t=t_N}_{t=0} \hat{\vec{E}}_{e}[t]   
\end{equation}

Recall that the GetSegments(.) subroutine in Alg.~\ref{alg:OCL} returns robust and non-robust segments after search on a robustness graph. The robustness metric for each edge is given by the algebraic computation in Eq.~\ref{eq:algebraic_computation}. 
\section{Robust planning and Control}
\label{sec:robust_planning}

We generate robust plans through trajectory optimization of a robust objective function $J_{R}$.

\begin{equation}\label{eq:objective_cp}
 J_{R}(\vec{X}, \vec{U}) = v_{e} {\hat{E}^{r^{2}}_{e}} + v_{m} {\hat{E}^{r^{2}}_{m}} + v_{j} J
\end{equation}
where $v_e$, $v_m$, and $v_{j}$ are positive constant weights.

To achieve robustness to state uncertainty under real-world model inaccuracies, we  minimize  $\hat{E}^{r}_{e}$ and $\hat{E}^{r}_{m}$,  the real expected and maximal divergence metrics respectively, for a given initial state and control sequence. 

$J$ is a deterministic objective for the task of reaching in clutter. 
%
%
Formally, the objective is a sum of running costs $L_{t}$ and terminal cost $L_{t_{N}}$ along a trajectory:
\begin{equation}\label{eq:objective}
 J (\vec{X}, \vec{U}) = v_{f} L_{t_{N}}(\vec{x}_{t_{N}}) + \sum^{t_{N-1}}_{t=t_{0}} L_{t}  
\end{equation}
where $v_{f} > 0$ is a constant weight. Details of $L_{t}$ and $L_{t_{N}}$ are given in Sec.~\ref{sec:cost_functions}.

We directly minimize the robust objective $J_R$ such that:
\begin{equation}\label{eq:traj_opt_problem}
    \vec{U}^{*} = \argmin_{\vec{U}} J_{R}(\vec{X}, \vec{U}) 
\end{equation}
subject to the dynamics constraint, $ \vec{x}_{t+1} = f(\vec{x}_{t}, \vec{u}_{t}) $, the terminal set constraint, $\vec{x}_{t_N} \in \mathbb{X}_{f}$, and the control sequence constraint, $\vec{U} \in \mathbb{U}$. 

The set of feasible terminal states, $\mathbb{X}_{f}$ is an $\alpha \geq 0$ sub-level set of the terminal cost function. Specifically:
\begin{equation}\label{eq:terminal_set}
    \mathbb{X}_{f} := \{ \vec{x} \hspace{1mm} | \hspace{1mm} L_{t_{N}}  \leq \alpha \}
\end{equation}
and the set of control-limited sequences is:
\begin{equation}\label{eq:feasible_controls_set}
    \mathbb{U} := \{ \vec{U} \hspace{1mm} | \hspace{1mm} \vec{b}_{l} \leq \vec{u}_{t} \leq \vec{b}_{u} \} 
\end{equation}
where $\vec{b}_{l}$ and $\vec{b}_{u}$ are constant vectors of lower and upper bound on the controls. 

A trajectory is feasible (in the deterministic sense) if it satisfies the terminal state constraint, the control limits, and yields a total cost $J$ less than a threshold $\beta > 0$. The cost threshold depends on the task. For reaching in clutter, it is defined such that no failures occur. Specifically, $\beta$ is selected such that no objects are toppled/dropped from the working surface, and there are no robot collisions with static obstacles, at the end of plan execution. 

\subsection{Robust sampling-based trajectory optimization}
\setlength{\textfloatsep}{2mm}
\begin{algorithm}[t]
    \SetKwInOut{Input}{Input}
    \SetKwInOut{Output}{Output}
    \SetKwInOut{Parameters}{Parameters}
    \SetKwInOut{Subroutines}{Subroutines}

    \Input{$\vec{x}_{0}$: Initial state \\ 
     $\vec{U}$: Initial candidate controls}
    \Output{$\vec{U}^{*}$: Feasible and robust control sequence \\
    $\vec{X}^{*}$: Feasible state sequence \\
    $\hat{\vec{E}}^{r}_{e}$: Real-world expected divergence metric\\
    }
    \Parameters{
     $S$: Number of noisy trajectory roll-outs \\
     $\bm{\nu}$: Sampling variance vector \\
      $I_{max}$: Maximum number of iterations}
  	 $\vec{X} , J \gets $TrajRollout$(\vec{x}_{0},\vec{U})$, \hspace{2mm} $I \gets 0$ \\
  	 $\hat{\vec{E}}^{r}_{e}$,  $\hat{\vec{E}}^{r}_{m}$  $\gets$  ComputeMetrics($\vec{X}$, $\vec{U}$) \\
  	  $J_{R} \gets $ Compute robust cost using Eq.~\ref{eq:objective_cp}\\
     \While{{$ I \leq I_{max}$} \textbf{and} ({$U$ not feasible  or $\hat{E}^{r}_{e} > 1 $}) }{
     \For{$s \gets 0 $ \KwTo $S-1$ }
     {
     $\vec{\delta{U}}^{s} \gets N(\vec{0},\bm{\nu})$  \\
     $\vec{U}^{s} = \vec{U} + \vec{\delta{U}}^{s}$ \\ 
     $\vec{X}^{s} , J^{s} \gets $TrajRollout$(\vec{x}_{0},\vec{U}^{s})$\\
       	 $\hat{\vec{E}}^{r_{s}}_{e}$,  $\hat{\vec{E}}^{r_{s}}_{m}$  $\gets$  ComputeMetrics($\vec{X}^{s}$, $\vec{U}^{s}$) \\
       	 $J^{s}_{R} \gets $ Compute robust cost using Eq. \ref{eq:objective_cp}
      }
      $s^{*}$ $\gets$ MinCostSample ($J^{0}_{R}$, $J^{1}_{R}$, \dots, $J^{S-1}_{R}$)\\ 
     \If {$J^{s^{*}}_{R} < J_{R}$ } 
     {
         $\vec{U} \gets \vec{U}^{s^{*}}$,
         $\vec{X} \gets \vec{X}^{s^{*}}$,
         $J_{R} \gets J^{s^{*}}_{R}$, 
         $\hat{\vec{E}}^{r}_{e}  \gets  \hat{\vec{E}}^{r_{s^{*}}}_{e}$ \\ 
     }
     $I \gets I+1$
     }
\Return   $\vec{U}$, $\vec{X}$, $\hat{\vec{E}}^{r}_{e}$
    \caption{Robust Stoch. Traj. Opt. (RobustSTO)}\label{alg:robust_sto}
\end{algorithm}
\setlength{\floatsep}{2mm}

Trajectory optimization methods such as STOMP \citep{stomp} have shown impressive speed for motion planning with parallel rollouts on multiple cores of a PC. They also easily accept arbitrary cost functions that may not be differentiable. 

In Alg.~\ref{alg:robust_sto}, we propose a robust sampling-based trajectory optimization algorithm (RobustSTO) for physics-based manipulation. It begins with an initial candidate control sequence $\vec{U}$ and seeks lower cost trajectories (lines 4-15) iteratively until a feasible and robust control sequence is found or the maximum number of iterations is reached (line 4). We add random control sequence variations $\vec{\delta U}^{s}$ on the candidate control sequence to generate $S$ new control sequences at each iteration (line 7). Thereafter on line 8, we roll-out each sample control sequence and return the corresponding state sequence $\vec{X}$ and cost $J$. A roll-out starts from the initial state $\vec{x}_{0}$ and applies each control input in $\vec{U}$ sequentially, one after the other. 

On line 9, we compute the maximal and expected divergence metric vectors along a given sample trajectory with the ComputeMetrics() subroutine. It involves $N_c$ trajectory roll-outs for each of the $N_w$ real-world realizations and also for the nominal model $f$. Then, we compute the robust cost of a sample trajectory using Eq.~\ref{eq:objective_cp}, on line 10. 

We take a greedy approach to trajectory updates. The minimum cost trajectory is selected as the update (line 11). However, the update is accepted only if it provides a lower cost (lines 12-13).  Finally, the algorithm returns feasible and robust controls (U), the corresponding nominal state sequence, $\vec{X}$, and the corresponding real-world expected divergence metric, $\hat{\vec{E}}^{r}_{e}$ (line 15).  

\subsection{Cost functions}
\label{sec:cost_functions}
For physics-based manipulation in clutter, we provide details of the objective. Let's begin with the running cost: 
\begin{equation}\label{eq:running_cost}
    L_{t}(\vec{x}_{t}, \vec{x}_{t+1}, \vec{u}_{t-1}, \vec{u}_{t}) = \sum_{i} v_{i} c_{i}, \hspace{1mm} i \in \{a,c,d,y\}
\end{equation}
where $v_{i} \geq 0$ is a constant weight, the cost terms penalize robot acceleration ($c_{a}$), scene disturbance ($c_{d}$), collision ($c_{c}$), and toppling ($c_{y}$). 
\subsubsection{Acceleration cost ($c_a$):}
We are interested in robot motion with minimal changes in robot velocity in between timesteps. We encode this desired behaviour in the robot acceleration cost:
\begin{equation}
    c_{a} (\vec{u}_{t-1}, \vec{u}_{t}) = || \vec{u}_{t} - \vec{u}_{t-1} ||^{2}
\end{equation}
\subsubsection{Disturbance cost ($c_d$):}
The disturbance cost penalizes displacing each of the $D$ dynamic objects from their initial positions and velocities:
\begin{equation}
    c_{d} (\vec{x}_{t}, \vec{x}_{t+1}) = \sum^{D}_{i=1} || \vec{x}_{t+1} - \vec{x}_{t} ||^{2}
\end{equation}
%


\subsubsection{Collision cost ($c_c$):} Here, we penalize collisions between the robot and all static objects in the environment through a discontinuous cost:
\begin{equation}
    c_{c} (\vec{x}_{t+1}) = 
    \begin{cases} 
      1 & $if robot collides with static objects$ \\
      0 & $otherwise$
   \end{cases}
\end{equation}
\subsubsection{Toppling cost ($c_{y}$):} We penalize toppling of objects during contact interaction through a discontinuous cost term:
\begin{equation}
    c_{y} (\vec{x}_{t+1}) = \sum^{D}_{i=1} C^{i}_{Topple}
\end{equation}
\begin{equation}
    C^{i}_{Topple} = 
    \begin{cases}
    1 & $if object i is toppled $ \\
    0 & $otherwise$
    \end{cases}
\end{equation}

In addition to the running cost, We also define the goal cost term for reaching in clutter as: 
\begin{equation}
    c_{g} (\vec{x}_{N}) = || \vec{r}_{GO} ||^2 + w_{\phi}\phi^{2}_{N}
\end{equation}
\begin{equation}
    \phi_{N} = \arccos (\hat{\vec{v}}_{G} \cdot \hat{\vec{r}}_{GO}) 
\end{equation}
where $\vec{r}_{GO}$ is the position vector from a point $G$ inside the gripper to a point $O$, the object's center of mass, $\hat{\vec{v}}_{G}$ is the gripper's forward unit direction, $w_{\phi} > 0$ is a constant weight, and $\phi_{N}$ is the angular distance between the gripper's forward unit direction and the unit vector $\hat{\vec{r}}_{GO}$. 

In some cases, a desired terminal state $\vec{x}^{d}$ can be provided to the optimizer. Then, we write the overall final cost function as:
\begin{equation}
    L_{t_{N}} (\vec{x}) = 
    \begin{cases}
    || \vec{x}^{d} - \vec{x} ||^{2} & $ if $\vec{x}^{d}$ is given$ \\
    c_{g}(\vec{x}) & $otherwise$
    \end{cases}
\end{equation}

A sampling-based, derivative-free approach is well suited to handle these discontinuous cost functions. 

\subsection{Time complexity of robust planner}
\label{sec:time_complexity}
Physics simulation is the most computationally expensive operation for the RobustSTO algorithm. This is done both in the TrajRollout(.) and in the ComputeMetrics(.) subroutines. 

Let $T_{u}$ be the physics simulation time for a single control input $\vec{u}_{t}$ applied for control duration $\Delta_{t}$. Then, the serial optimization time,
$T^{s}$ is a sum of $T_{u}N ($line 1$) $, $T_{u}NN_{c}(N_w + 1)$ (line 2), $T_{u}NSI$ (line 8), and $T_{u}NN_{c}(N_w + 1)SI$ (line 9). 

Thus, the serial computation time is:

\begin{align}
    T^{s} = T_{u}N(SI+1)(N_{c}(N_{w} + 1) + 1)
\end{align}

Rollouts either in TrajRollout(.) or in ComputeMetrics(.) can be computed simultaneously in parallel. Thus, a parallel implementation gives an optimization time $T^{p}$, independent of $S$, $N_w$, and $N_c$:

\begin{align}
    T^{p} = T_{u}N(I + 1)
\end{align}

Hence, with sufficient parallel cores available ($N_c$ $\times$ $N_w$ $\times$ $S$), we find that RobustSTO is a polynomial time algorithm with complexity $\mathcal{O}(N(I+1))$.   

\subsection{Model-Predictive Control}
\label{sec:mpc}
We use a model-predictive controller (MPC) as the closed-loop controller in this work. It calls the trajectory optimizer online, during execution, to solve a finite horizon optimal control problem. It executes the first action in the control sequence, updates the internal state with camera feedback, and then runs the trajectory optimizer again to generate a new control sequence, and repeats this process at every step. We use only the deterministic objective $J$ during MPC, i.e, a version of Alg.~\ref{alg:robust_sto} without robustness computations. 

\begin{figure*}
	\begin{subfigure}[b]{0.33\textwidth}
	\centering
		\includegraphics[height=1.7in, width=2.2in, angle=0]{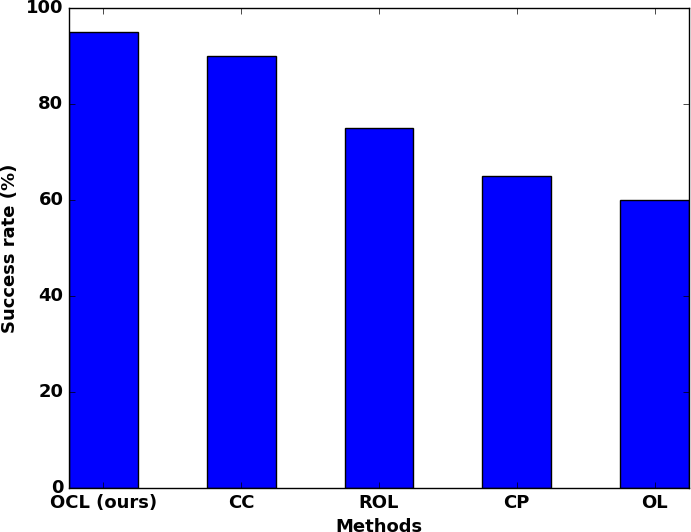}
		\caption{Success rate}
		\label{fig:real_world_execution_success}
	\end{subfigure}
	\begin{subfigure}[b]{0.33\textwidth}
	 \centering 
		\includegraphics[height=1.7in, width=2.2in, angle=0]{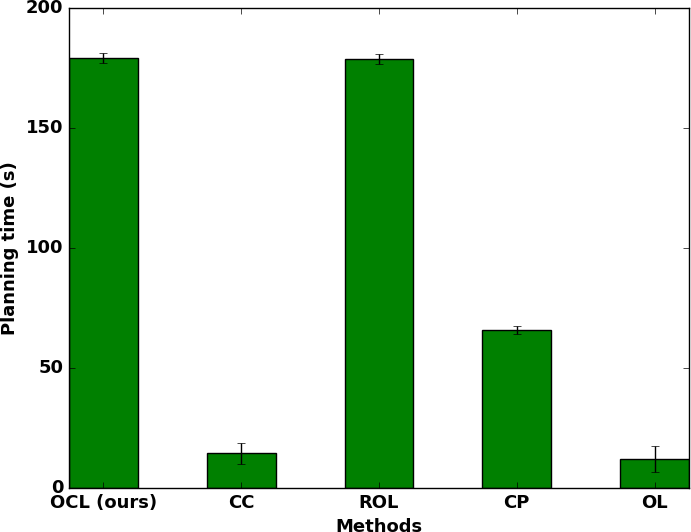}
		\caption{Planning time}
		\label{fig:planning_time}
	\end{subfigure}
	\begin{subfigure}[b]{0.33\textwidth}
	\centering 
		\includegraphics[height=1.7in, width=2.2in, angle=0]{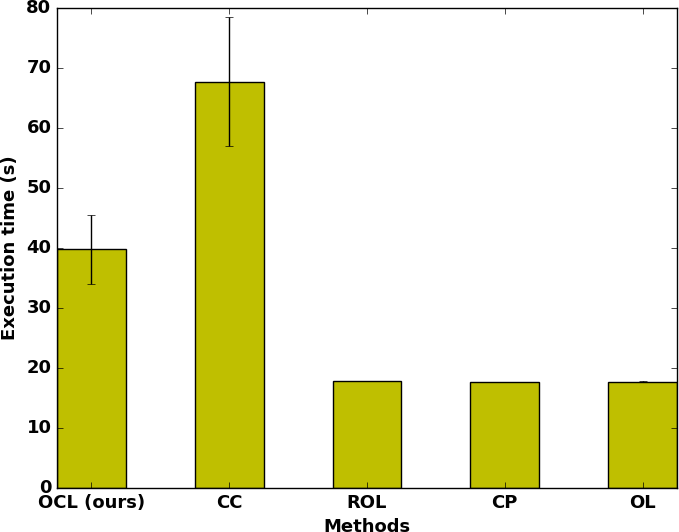}
		\caption{Execution time}
		\label{fig:execution_time}
	\end{subfigure}
	\caption{
	Experimental results comparing the interleaved open and closed-loop execution method ($OCL$) with other baselines. We randomly created 20 real-world scenes and tested all five methods on each scene. Here we show success rate, planning time, and execution time plots. All error bars indicate a 95\% confidence interval of the mean. Partially open-loop plans from $OCL$ achieve a similar success rate to closed-loop control, while spending less time on execution. This comes at a price of increased initial planning time. However, the planning time for $OCL$ (and $ROL$) can be made similar to that of $CC$ with more parallel cores (Sec.~\ref{sec:time_complexity}).}
	\label{fig:real_world_exp_results}
\end{figure*}
\section{Robot Experiments and Results}
\label{sec:experiments}

Through our experiments, we ask two important questions:

\begin{itemize}
    \item How does the interleaved open and closed-loop execution planning and control framework compare with only open-loop and only closed-loop approaches?
    \item How does the normal divergence metric proposed in prior work compare with the real-world divergence metric proposed in this paper, for manipulation in the real-world? 
\end{itemize}

We answer these questions through real-world robot experiments.

\subsection{Setup} 
\label{subsec:setup_exp}
The robot is a 6 degrees of freedom (DOF) UR5 arm with a 1-DOF Robotiq 2-finger gripper. We use Mujoco\cite{mujoco} as the physics simulator to plan all robot actions. The planning environment consists of a shelf and 10 different objects, including some from the YCB dataset \cite{Calli_RAM_2015}. Object pose is captured with the OptiTrack motion capture system. In Table~\ref{table:parameters}, we detail all parameters used throughout the experiments.
\begin{center}
\begin{table}[t]
\caption{Experimental parameters} 
\centering 
\begin{tabular}{c c} 
\hline\hline 
 & Parameter value \\ 
\hline \hline 
\textbf{Robustness graph}\\
\hline\hline
[$C_{ro}$, $C_{nr}$ ] & [1, 1000] \\
\hline\hline
\textbf{Divergence metrics}\\
\hline\hline
$[N_c, N_w]$ & [4, 4]\\
$[m_{l}, m_{u}]$ & $[0.5, 0.8] kg$ \\
$[\mu_{l}, \mu_{u}]$  & $[0.2, 0.4]$ \\
\hline\hline
\textbf{Robust planning}\\
\hline\hline
$[\alpha, \beta ]$ & [10, 50] \\
$[S, N, \Delta_{t}]$  & [4, 5, 0.2s] \\
$[\bm{b}_{l}, \bm{b}_{u}]$ & $\pm$ $\pi$~rad/s  \\
$I_{max}$ & 10 \\
$[v_{e}, v_{m}, v_{j}]$  & [2, 0.5, 1]\\
$[v_{a}, v_{c}, v_{d}, v_{y}, v_{\phi}, v_{f}]$ & [0.001, 200, 1000, 200, 0.019]\\
\hline\hline
\textbf{Environment}\\
\hline\hline
D & 10 \\
m & 7 \\
\hline\hline
%
\end{tabular}
\label{table:parameters}
\end{table}
\end{center}

\subsection{Baselines} 
In this work, we compare against four baselines:

\subsubsection{Open-loop execution (OL)}
This is an approach where we minimize the deterministic objective J with trajectory optimization. It produces a sequence of controls that are then executed open-loop in the real-world. 

\subsubsection{Robust open-loop execution (ROL)}
This baseline minimizes the robust objective with trajectory optimization. The trajectory is then executed open-loop in the real-world. 

\subsubsection{Convergent planning (CP)} 
This baseline uses divergence metrics in \citet{convergent_planning} to generate robust plans through trajectory optimization, and executes them open-loop. Specifically, it uses only the "normal" divergence metrics $\hat{E}^{f}_{e}$ for planning, without the real-world metric computations considered in this work. 

\subsubsection{Closed-loop control (CC)}
The closed-loop control approach performs trajectory optimization at every step, during execution, using the deterministic objective J and then executes only the first planned control. It re-plans online until task completion. 

\subsection{Comparison of OCL with baselines}
We generate 20 real-world scenes (see Fig.~\ref{fig:sample_plans}), by placing the target object behind other objects, such that reaching directly for it is almost impossible. We also pick different objects to surround the target in each scene. We fix the robot's initial configuration in all scenes to allow for easy scene reset. We run each baseline and $OCL$ to complete the manipulation task. That's a total of 100 robot manipulation runs - 5 methods per scene. We recorded success, planning time, execution time, divergence metrics, and finally what percentage of trajectory segments are executed open-loop vs. closed-loop in a given scene (for $OCL$). 

Results are shown in Fig.~\ref{fig:real_world_exp_results}, Fig.~\ref{fig:real_or_normal}, and Fig.~\ref{fig:percentage_executions}.  

\subsection{Execution success}
We find that the proposed method, $OCL$, is \textit{more successful} in completing the reaching in clutter tasks, compared to all the other baselines. As shown in Fig.~\ref{fig:real_world_execution_success}, open-loop execution ($OL$) has the least success rate - a difference of about 35\% compared to the \textit{interleaved open and closed-loop} approach proposed in this work. An interesting observation is that the interleaved open and closed-loop approach has a similar success rate to closed-loop control, which is impressive considering the fact that $OCL$ contains partially open-loop plans. 

\subsection{Planning time}
The planning time is the total time spent by planners before the robot executes its first action. 

We show the planning time in Fig.~\ref{fig:planning_time}.

On a 4 core PC, the robust planning time is higher than that of planning without robustness metrics ($OL$ and $CC$). However, the \textit{estimated} robust planning time on a PC with 64 ($N_{c} \times N_{w} \times S$) cores, would be similar to standard planning ($OL$ and $CC$). This is due to easy parallelization of trajectory roll-outs (Sec.~\ref{sec:time_complexity}), provided the cores are available.  

\subsection{Execution time}

The execution time is the total time spent after the robot starts executing it's first action. 

As shown in Fig.~\ref{fig:execution_time}, the execution time of $OCL$ is close to that of traditional open-loop execution (i.e $ROL$, $OL$, and $CP$) but much lower than full closed-loop control. 

This difference would be \textit{much higher} for scenes with less number of objects, since completely open-loop plans could be found. However, even in such simple scenes, a fully closed-loop approach would stop at each step to re-plan, taking much more time to complete the manipulation task.  

\begin{figure}[t]
    \begin{subfigure}[b]{0.24\textwidth}
    \centering
     \includegraphics[scale=0.35]{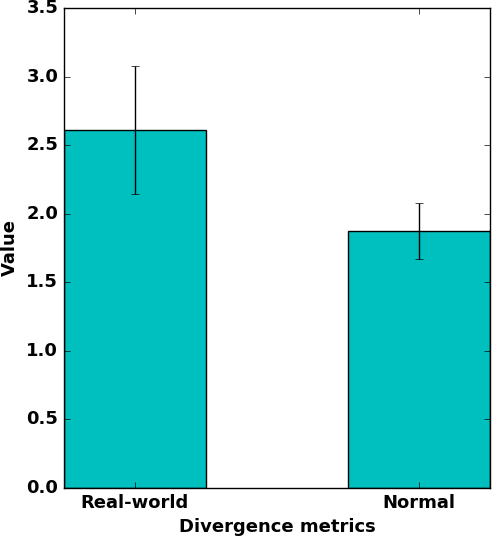}
     \subcaption{Divergence metrics}
     \label{fig:real_or_normal}
    \end{subfigure}
    \begin{subfigure}[b]{0.24\textwidth}
    \centering
    \includegraphics[scale=0.51]{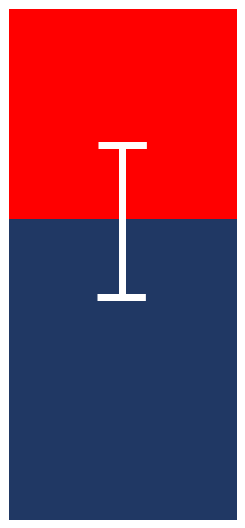}
    \begin{picture}(0,0)
    \put(-58.5, 122){\textcolor{white}{Open-Loop}}
    \put(-48, 110){\textcolor{white}{(38\%)}}
    \put(-61, 30){\textcolor{white}{Closed-Loop}}
    \put(-48, 15){\textcolor{white}{(62\%)}}
    \end{picture}
    \subcaption{Open or closed-loop actions}
    \label{fig:percentage_executions}
    \end{subfigure}
		\caption{
		(a) Divergence metrics recorded from real-robot experiments for open-loop methods. Real-world divergence is recorded for $ROL$. Normal divergence is recorded for $CP$. We find that the real-world divergence metric was higher than normal divergence metrics in prior work, yet $ROL$ proved to be more successful (Fig.~\ref{fig:real_world_execution_success}). Results are shown within 95\% confidence interval of the mean.
		(b) Percentages of open and closed-loop actions in a given manipulation plan. On average 38\% of planned actions in a given trajectory were executed open-loop, while the remaining 62\% of actions were executed closed-loop. The white bar shows the 95\% confidence interval of the mean which is 13.81\%.}
\end{figure}
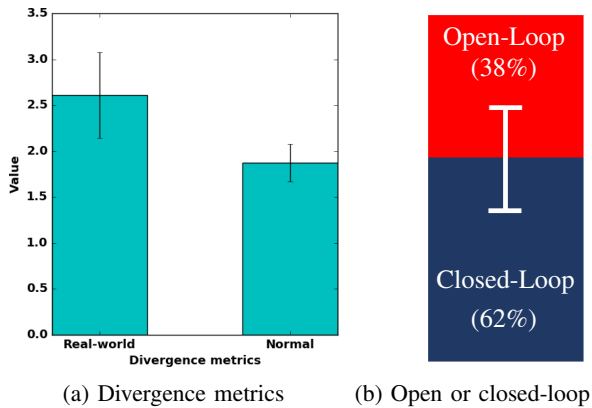


\subsection{Real or normal divergence metrics}
Normal divergence metrics proposed in prior work do not consider model inaccuracies. How does the real divergence metric compare with the normal divergence metric, for robust planning, and execution in the real-world? Based on data from the baseline experiments, we attempt to answer this question. 

We found that the expected normal divergence metrics ($CP$) are much lower than the expected real divergence metrics ($ROL$), as shown in Fig.~\ref{fig:real_or_normal}. Implying that $CP$ should be more successful than $ROL$. However, this is not the case. We've seen previously in Fig.~\ref{fig:real_world_execution_success} that $ROL$ is more successful than $CP$ by about 10\%. Thus, the real-world divergence metric is a better estimate of the uncertainty for the reaching in clutter task. 

\subsection{Composition of $OCL$ plans}
\begin{figure*}[htb!]
    \centering
    \vspace{5mm}
    \includegraphics[scale=0.87]{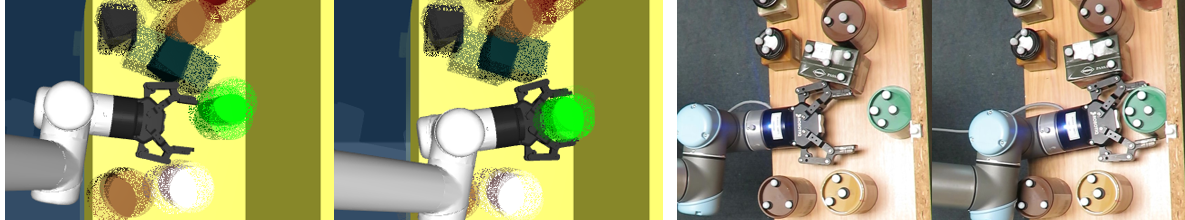}
    \begin{picture}(0,0)
        \put(-170,120){\textbf{Robustness computation}}
        \put(120, 120){\textbf{Real-world execution}}
    \end{picture}
    \vspace{-3mm}
    \caption{A sample robust segment from the interleaved open and closed-loop approach. The left image shows a part of the robustness metric computation. We see that the initial state uncertainty is reduced through a funnelling action that pushes the target object (green cylinder) towards the shelf, and grasps it. On the right image we see the robust segment executed open-loop in the real world.}
    \label{fig:metrics_scene_3}
\end{figure*}
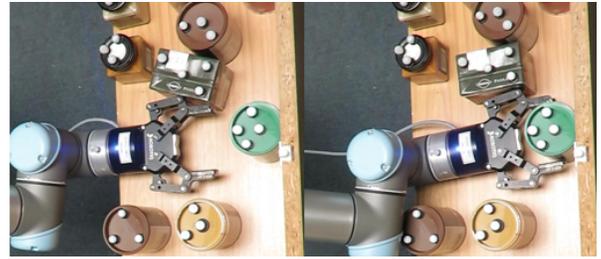

\begin{figure*}[htb!]
    \centering
    \vspace{5mm}
    \includegraphics[scale=0.87]{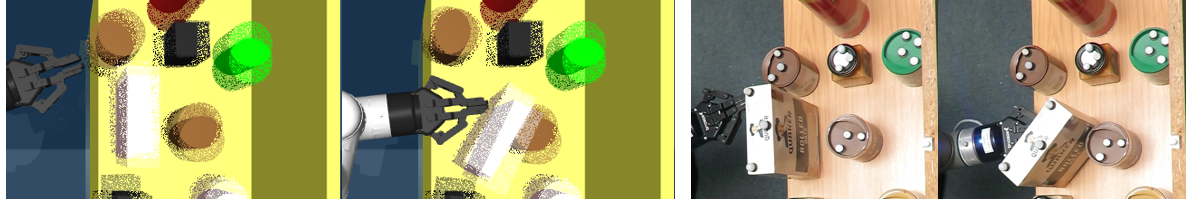}
    \begin{picture}(0,0)
        \put(-170,110){\textbf{Robustness computation}}
        \put(120, 110){\textbf{Real-world execution}}
    \end{picture}
    \vspace{-3mm}
    \caption{A robust segment from $OCL$. The robustness metric computation is shown on the left image where the initial state uncertainty is reduced through a stable side push. On the right image we see the robust segment executed open-loop in the real world.
    }
    \label{fig:metrics_scene_1}
\end{figure*}
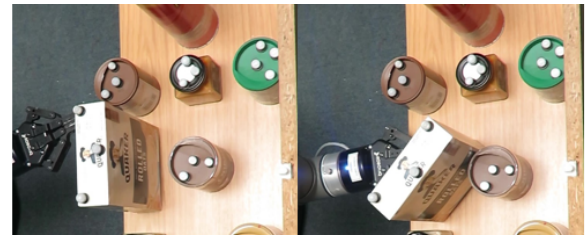

Using the interleaved open and closed-loop approach, what percentage of a trajectory is executed open-loop versus closed-loop? 

We recorded the percentages during $OCL$ experiments. Results are shown in Fig.~\ref{fig:percentage_executions}. We found that 62\% of a plan was executed closed-loop, while 38\% was executed open-loop, on average.  

Note that these percentages will change depending on task difficulty, especially the number of contact interactions. For example, one might see a higher percentage of open-loop execution for a scene with only a few number of objects. 

In this way, the interleaved open and closed-loop approach ($OCL$) has the potential to adapt to varying task difficulty, executing fully open-loop or fully closed-loop or anything in-between. 

\subsection{Examples of robust trajectory segments} 

What sort of robust trajectory segments were encountered during experiments? 

One major robustness strategy is the funnelling action, where the robot pushes an object towards the shelf wall and grasps it. This is shown in Fig.~\ref{fig:metrics_scene_3}. On the left image, we show part of the robustness metric computation where initial state uncertainty is reduced, through a robust robot motion. On the right, we see the real-world execution during one of our experiments. 

Another robustness strategy is a stable side push on an object as shown in Fig.~\ref{fig:metrics_scene_1}. The box is pushed towards the cylinder in a robust way, where the initial state uncertainties are reduced.  

One final robustness strategy for the robot is to avoid any contact and move completely in free space. We've seen this strategy in Fig.~\ref{fig:robot_fig_1}.

\subsection{Sample plans from the different planners}

\begin{figure*}[htb!]
\centering
\begin{subfigure}[b]{\textwidth}
\centering 
\includegraphics[scale=0.8]{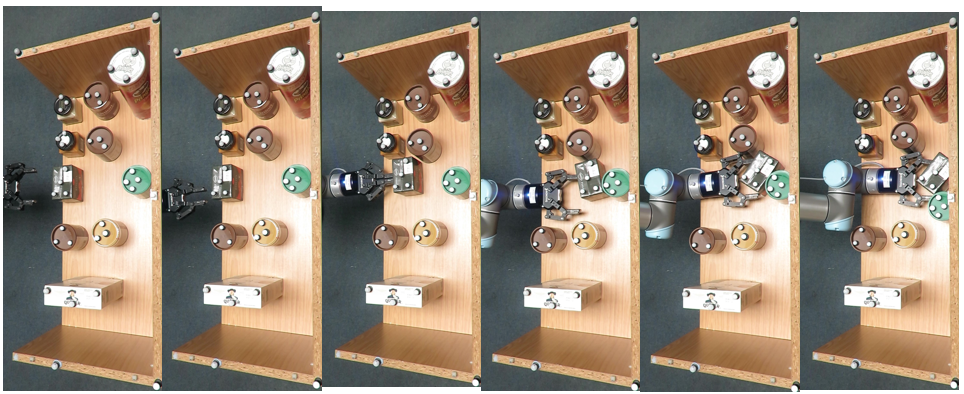}
\end{subfigure}
%
\begin{subfigure}[b]{\textwidth}
\centering
\includegraphics[scale=0.76]{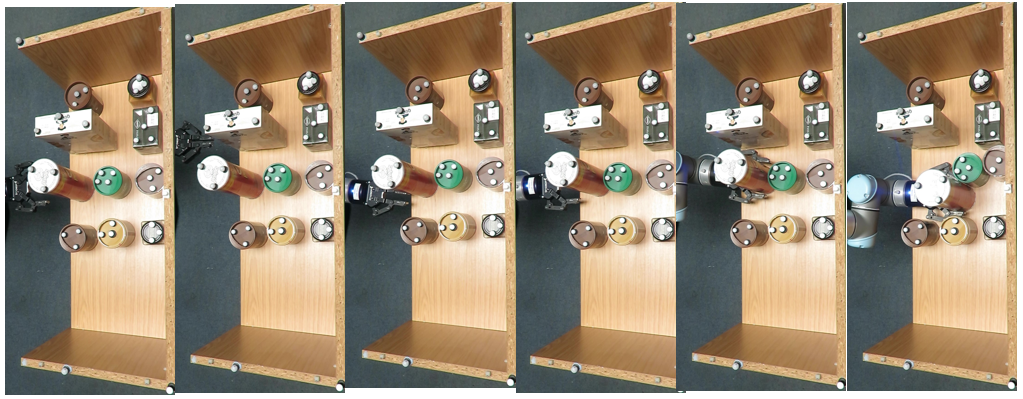}
\end{subfigure}

\vspace{-5mm}

\begin{subfigure}[b]{\textwidth}
\centering
\includegraphics[scale=0.81]{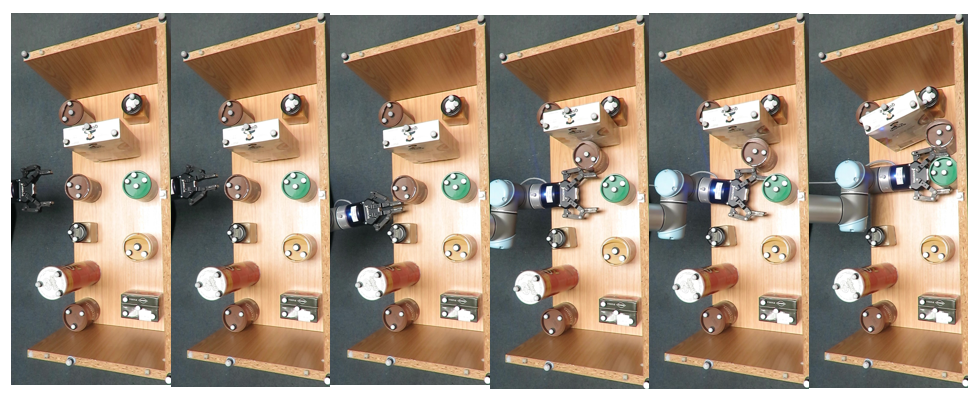}
\end{subfigure}
\vspace{-3mm}
\begin{subfigure}[b]{\textwidth}
\centering
\hspace{1mm}
\includegraphics[scale=0.76]{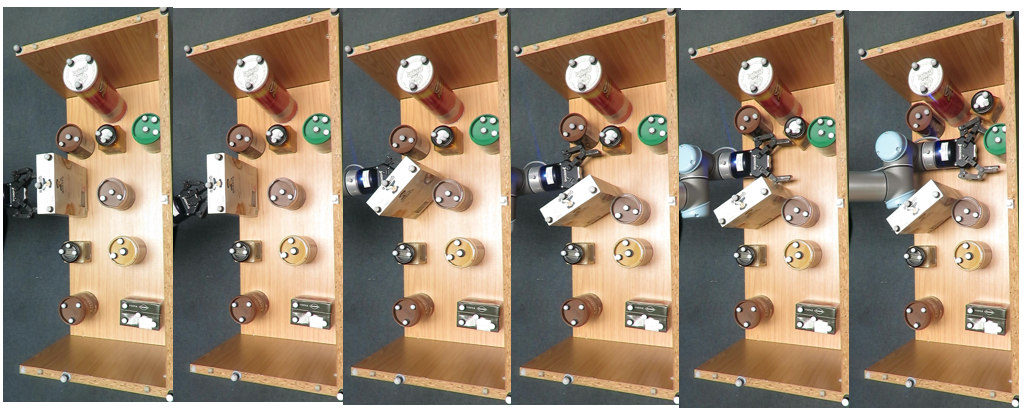}
\end{subfigure}
\begin{picture}(0,0)
\put(-210,510){\rotatebox{90}{ $OL$ failure}}
\put(-210,365){\rotatebox{90}{$CP$ failure}}
\put(-210,208){\rotatebox{90}{$ROL$ success}}
\put(-210, 60){\rotatebox{90}{$CC$ success}}
\end{picture}
\caption{Examples of successful and failed manipulation plans from different planning and execution methods. One major failure mode in these scenes is when the robot ends up with a different object in its grippers.}
\label{fig:sample_plans}
\end{figure*}

In Fig.~\ref{fig:sample_plans}, we show several plans from all baselines tested in the real-world. We show these in four different scenes.

In the first row, we see an open-loop execution ($OL$) failure. The black box didn't move out of the way as planned. It ended up in the gripper at the final state. This is a failure. 

In the second row we see a trajectory planned through the normal divergence metrics ($CP$). The motion of the cylinder is different from originally planned in simulation. Thus, the cylinder ended up in the robot's gripper - a failure case. 

In the third row, we show a successful run from the robust open-loop execution ($ROL$) method where we use the real-world divergence metrics proposed in this work. The robot pushes the brown cylinder in a robust manner onto other supporting objects. It successfully reaches for the green target object. 

Finally, in the last row, we show a closed-loop control run, where the robot re-plans at every step to successfully reach for the green target object. 

\section{Discussion and Future Work}
\label{sec:conclusion}
 \begin{figure}[t]
\centering 
  \begin{subfigure}[b]{0.24\textwidth}
  \centering
      \includegraphics[height=1.45in,width=1.75in]{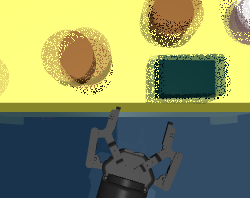}
  \end{subfigure}
  \hspace{-3mm}
  \begin{subfigure}[b]{0.24\textwidth}
  \centering 
   \includegraphics[height=1.45in,width=1.75in]{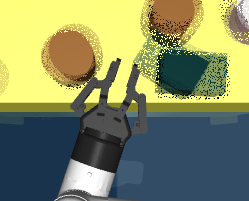}
   \end{subfigure}
  \caption{An example of robust and non-robust object motions in a given trajectory segment. On the left image, each object takes several possible positions. This illustrates the initial state uncertainty. After a robot motion that pushes on both a cylinder to the left and a box to the right, the size of the state uncertainties change. It is smaller for the cylinder to the left, and larger for the box to the right. Hence the cylinder's motion is robust and the boxes' motion is not. }
  \label{fig:object_motions}
 \end{figure}
 In this work, we present for the first time, an interleaved open and closed-loop control framework for physics-based manipulation under uncertainty. We derived robustness metrics through contraction theory, and used these metrics to plan robust robot motions. We separated a trajectory into robust and non-robust segments through a minimum cost search on a directed robustness graph. Robust segments are executed open-loop while non-robust segments are executed with model-predictive control. 
 We show through experiments on a real robotic system, that the interleaved open and closed-loop approach produces partially open-loop robust plans with success rates comparable to closed-loop control, while achieving a more fluent/real-time execution. 
 
 \subsection{Robust and non-robust object motions}
 In Fig.~\ref{fig:object_motions}, we show an example of object motions that result in an overall robust or non-robust trajectory. 

We generate several sample initial states as shown on the left image, varying initial object positions, around their current position. Then we apply the forward robot motion shown. In the right image we see the resulting state distribution for several objects, at the end of a single action. The object position distribution is smaller at the final state for the cylinder compared with the initial state. Hence, this is a robust action for the cylinder. However, the opposite is the case for the box. The final object position distribution is larger for the box. This is due to the fact that sometimes the gripper makes contact with it, and rotates it, and other times it does not. The action is non-robust for the box. 

The overall robustness of an action in these scenes is then judged through the aggregate of these robust and non-robust object motions. 

An important direction for future work is to identify 'important' objects of interest for a given task. Then one can focus planning on making these object motions robust. 
 
\bibliographystyle{unsrtnat}
\bibliography{new_bib} 

\end{document}